\documentclass{article}

\PassOptionsToPackage{numbers,compress}{natbib}
\PassOptionsToPackage{table,x11names}{xcolor}

\usepackage[preprint]{neurips_2022}

\usepackage[utf8]{inputenc} % allow utf-8 input
\usepackage[T1]{fontenc}    % use 8-bit T1 fonts
\usepackage{hyperref}       % hyperlinks
\usepackage{url}            % simple URL typesetting
\usepackage{booktabs}       % professional-quality tables
\usepackage{amsfonts}       % blackboard math symbols
\usepackage{nicefrac}       % compact symbols for 1/2, etc.
\usepackage{microtype}      % microtypography
\usepackage{xcolor}         % colors

\title{Deep Clustering with Features from Self-Supervised Pretraining }

\author{%
	Xingzhi Zhou\quad Nevin L. Zhang \\
	Department of Computer Science and Engineering\\
	The Hong Kong University of Science and Technology\\
	\texttt{\{xzhoubl, lzhang\}@cse.ust.hk} \\
}

\usepackage{booktabs}
\usepackage{subfigure}
\usepackage{float}
\usepackage{amsmath,amssymb,bm} % define this before the line numbering.
\usepackage{graphicx}
\renewcommand{\mathbf}[1]{\bm{#1}}

\usepackage{adjustbox}

\usepackage{array}
\newcolumntype{x}[1]{>{\centering\arraybackslash}p{#1}}

\usepackage[ruled,lined, linesnumbered]{algorithm2e}

\usepackage{array, booktabs, boldline} %

\usepackage{cellspace}

\begin{document}

\maketitle

\begin{abstract}
A deep clustering model conceptually consists of a feature extractor that maps data points to a latent space, and a clustering head that groups data points into clusters in the latent space. Although the two components used to be trained jointly in an end-to-end fashion, recent works have proved it beneficial to train them separately in two stages. In the first stage, the feature extractor is trained via self-supervised learning, which enables the preservation of the cluster structures among the data points. To preserve the cluster structures even better, we propose to replace the first stage with another model that is pretrained on a much larger dataset via self-supervised learning.  The method is simple and might suffer from domain shift. Nonetheless, we have empirically shown that it can achieve superior clustering performance. When a vision transformer (ViT) architecture is used for feature extraction, our method has achieved clustering accuracy 94.0\%, 55.6\% and 97.9\% on CIFAR-10, CIFAR-100 and STL-10 respectively. The corresponding previous state-of-the-art results are 84.3\%, 47.7\% and 80.8\%.
Our code will be available online with the publication of the paper.
	
	% v3.1 the abstract is updated on May 15th 18:00.
	
\end{abstract}

\section{Introduction}

Deep learning has achieved human-level performance on image classification \cite{he2015delving}. It would be academically interesting to develop deep learning algorithms that can match human’s ability to cluster images.  Such work also has practical significance because it can benefit data organization, data labeling, information retrieval, and other deep learning algorithms.

Deep clustering conceptually consists of two phases: (1) Map data points to a latent space, and (2) group data points into clusters in the latent space.  The success of deep clustering depends critically on whether the cluster structures among the data points are preserved in the first phase. By {\em cluster structure preservation} we mean that semantically similar images are placed close to each other in the latent space, and semantically dissimilar images are placed apart from each other. Cluster structure preservation is extremely difficult due to the flexibility of neural networks.

Early deep clustering methods train the two phases jointly in an {\em end-to-end (E2E)} fashion and place emphasis on clustering-friendly features. They use a variety of losses to encourage data points to form tightly packed and well-separated clusters in the latent space \cite{dilokthanakul2016deep, guo2017improved, jiang2017variational, xie2016unsupervised, yang2017towards}. However, no explicit efforts are made to ensure the clusters in the latent space correspond well to semantic clusters in the input space. There are methods that use locality-preserving  to explicitly encourage cluster structure preservation \cite{huang2014deep, chen2017unsupervised}. However, their success is limited due to the difficulty in defining semantically meaningful distance metrics in the image space.

We conceptualize a general strategy for cluster structure preservation in feature mapping. We metaphorically term it the {\em  black-sheep-among-white-sheep (BaW)} strategy.  Imagine several (relatively small) herds of black sheep on an open grass field. The different herds would easily mix up if the black sheep are by themselves. However, this is less likely to happen if a large number of white sheep are also present. In deep clustering, the data points to be clustered (the target data) are the black sheep and other auxiliary data points are the white sheep.
In the BaW strategy, one builds a feature mapping from the white sheep data using self-supervised learning, use it to map the  black sheep data points to a latent space, and cluster the data points there.

 The BaW strategy is used in several recent deep clustering algorithms SCAN \cite{van2020scan}, NNM \cite{Dang_2021_CVPR}, and TSUC \cite{hanMitigatingEmbeddingClass2020}.
Those methods consist of two stages. In the first stage, they obtain a feature mapping by applying self-supervised learning on the target (black sheep) data. Augmented data are produced in the process and they are the white sheep.  We call those {\em Two-stage deep clustering with Self-supervised Learning} or  {\em TSL} methods.

In this paper, we propose to
use another dataset from a separate source as the white sheep and train a feature map on them using self-supervised learning. The advantage of this method is the flexibility of using a potentially very large dataset for pretraining, a recipe that has been proven effective for supervised learning in both natural language processing \cite{devlin2018bert} and computer vision \cite{he2019rethinking}.
We call this the
 {\em Two-stage deep clustering with Self-supervised Pretraining} or    {\em TSP} method.

\begin{figure}[t]
	\centering
		\begin{minipage}{0.48\linewidth}
		\includegraphics[width=1 \linewidth]{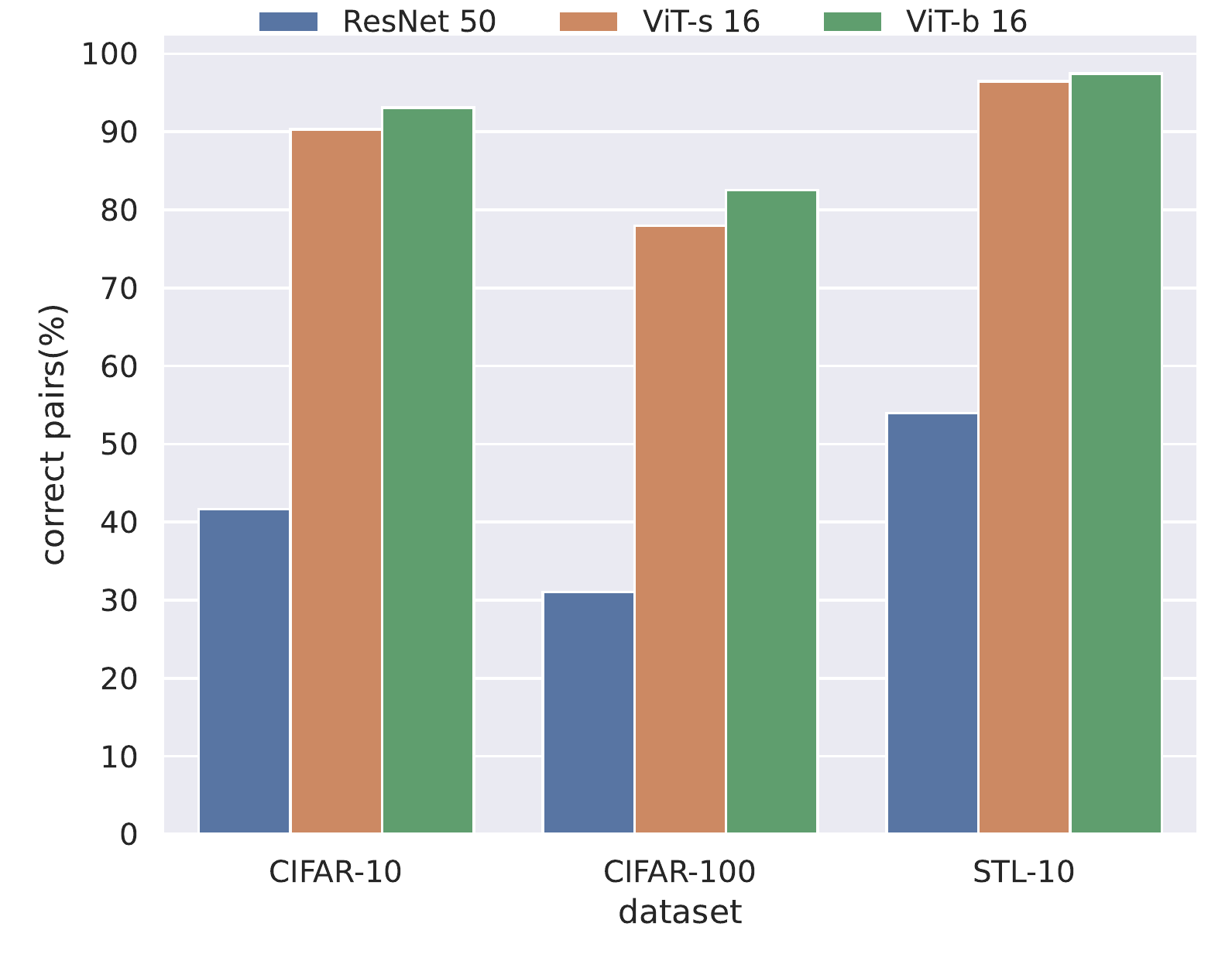}
		\caption{Label consistency of 20 nearest neighbors on CIFAR-10, CIFAR-100 and STL-10. The distance metric is the inner product on the representations obtained by different architectures pretrained on ImageNet-1k by DINO}
		\label{fig:consistency of nearest neighbors}
	\end{minipage}
\hfill
	\begin{minipage}{0.45 \linewidth}
		\includegraphics[width=1 \linewidth]{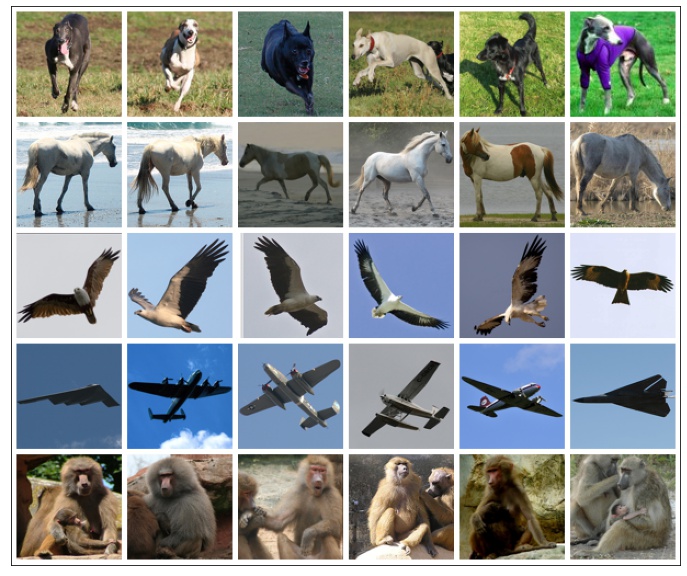}
			\caption{Images in the first column and five nearest neighbors in the other columns on the dataset STL-10. The distance metric is the inner product of representations on STL-10 obtained by ViT-s 16 pretrained on ImageNet-1k by DINO}
			\label{fig:nearest_neighbors_visualization}
	\end{minipage}

\end{figure}

The pretraining data and the target data typically have different distributions. In other words, domain shift is an issue. A key question is whether the cluster structures  can be preserved despite the domain shift. The answer turns out to depend on the choice of architecture for the feature extractor.  In our studies, we considered the use of architectures ResNet50 \cite{he2016deep}, ViT-s 16 and ViT-b 16 \cite{touvron2021training} for feature extraction, and pretrained them on ImageNet-1k \cite{russakovsky2015imagenet} using a recent self-supervised learning algorithm DINO \cite{caron2021emerging}. We then applied the feature extractors on CIFAR-10, CIFAR100 and STL-10, and we checked whether the 20 nearest neighbors of each data point in the latent  space share the same class label (and hence semantically similar in the image space) as a way to gauge if the cluster structures among data points are preserved.  The results are shown in Figure \ref{fig:consistency of nearest neighbors} and \ref{fig:nearest_neighbors_visualization}.  It is clear that the use of ViT architectures preserves the cluster structures well, which is consistent with previous observation on the relative robustness of ViTs against domain shift  \cite{naseer2021intriguing}. However, the same is not true for ResNet50.

Based on the above findings, we developed a TSP method for deep clustering.
We use a ViT-architecture for feature extraction
and a loss function similar to SCAN \cite{van2020scan} for the clustering head.
 Unlike previous TSL methods, our TSP method keeps the feature extractor frozen in the second stage.
The reason is that fine-tuning the feature extractor might negatively impact cluster structure preservation because it involves only the black sheep.
We tested TSP-ViT on CIFAR-10, CIFAR-100 and STL-10, and achieved clustering accuracy 94.0\%, 55.6\% and 97.9\% respectively. Those are drastically better than the corresponding previous state-of-the-art results 84.3\%, 47.7\%, and 80.8\%.  In summary, the contributions of this papers are as follows:

\begin{itemize}
	\item We highlight the importance for the feature mapping in deep clustering to be cluster structure preserving, and propose BaW as a general strategy to achieve cluster structure preservation.
	
	\item	We propose a novel two-stage deep clustering method TSP where a pretrained model is used for feature extraction.  While the use of self-supervised pretraining in supervised learning is now commonplace, we are unaware of any previous work that uses self-supervised pretraining for clustering.  In fact, we believe that such a method would be premature if proposed earlier because only recently we have models that are relatively robust to domain shift.
	
	\item We empirically show that our TSP method, when coupled with ViT, drastically improves the state-of-the-art in deep clustering.
\end{itemize}

\section{Related works}
% finish update on May 15 22:22

\textbf{Deep Image clustering}
We will use several previous clustering methods as baselines in our experiments.
Among them, DAC \cite{chang2017deep} and DCCM \cite{wu2019deep} are adaptive methods that
 pick positive and negative pairs based on the current model parameters to improve the model parameters further.
  IIC \cite{ji2019invariant} is conceptually similar to contrastive learning
  and aims to assign  two augmented versions of an image to the same cluster.
 PICA \cite{huang2020deep} aims to maximize the margins between different clusters.
 
It is argued in
 \cite{van2020scan} and  \cite{hanMitigatingEmbeddingClass2020}
 that
  end-to-end clustering methods such those mentioned above 
  depend heavily on model initialization and are likely  to latch onto low-level features such as color, texture and pixel intensity.
   DEC \cite{xie2016unsupervised} is a two-stage method where the first stage 
   initializes a feature mapping using denoising autoencoders. It suffers from 
   similar shortcomings. 
   To overcoming those shortcomings, SCAN \cite{van2020scan} and  TSUC \cite{hanMitigatingEmbeddingClass2020} are independently proposed.  They use self-supervised learning to learn the initial feature mapping. NNM \cite{Dang_2021_CVPR}  is an improvement of SCAN.
  In this paper, we view the difference between those three recent methods and earlier methods from  another perspective, i.e., whether ``white sheep" are used to separate ``black sheep",    and propose to learning a feature mapping via self-supervised pretraining on ``white sheep" from a separate source.

\textbf{Self-supervised learning}  Self-supervised learning methods learn data representations  using pretext tasks such as rotation \cite{gidaris2018unsupervised}, jigsaw puzzle \cite{noroozi2016unsupervised}, clustering \cite{caron2018deep}, contrastive learning \cite{chen2020big, caron2020unsupervised}, and so on. In particular,  BYOL \cite{grill2020bootstrap} learns data representation by matching features from a momentum encoder with those for different augmentations of one image.  DINO \cite{caron2021emerging} applies the idea of BYOL to vision transformer (ViT), and adds the multi-crop technique from \cite{caron2020unsupervised}.  It is one of the state-of-the-art self-supervised learning methods.

\section{Methodology}
%Copied from the version from the professor%

\begin{figure}[t]
	\centering
	\includegraphics[width=0.85\linewidth]{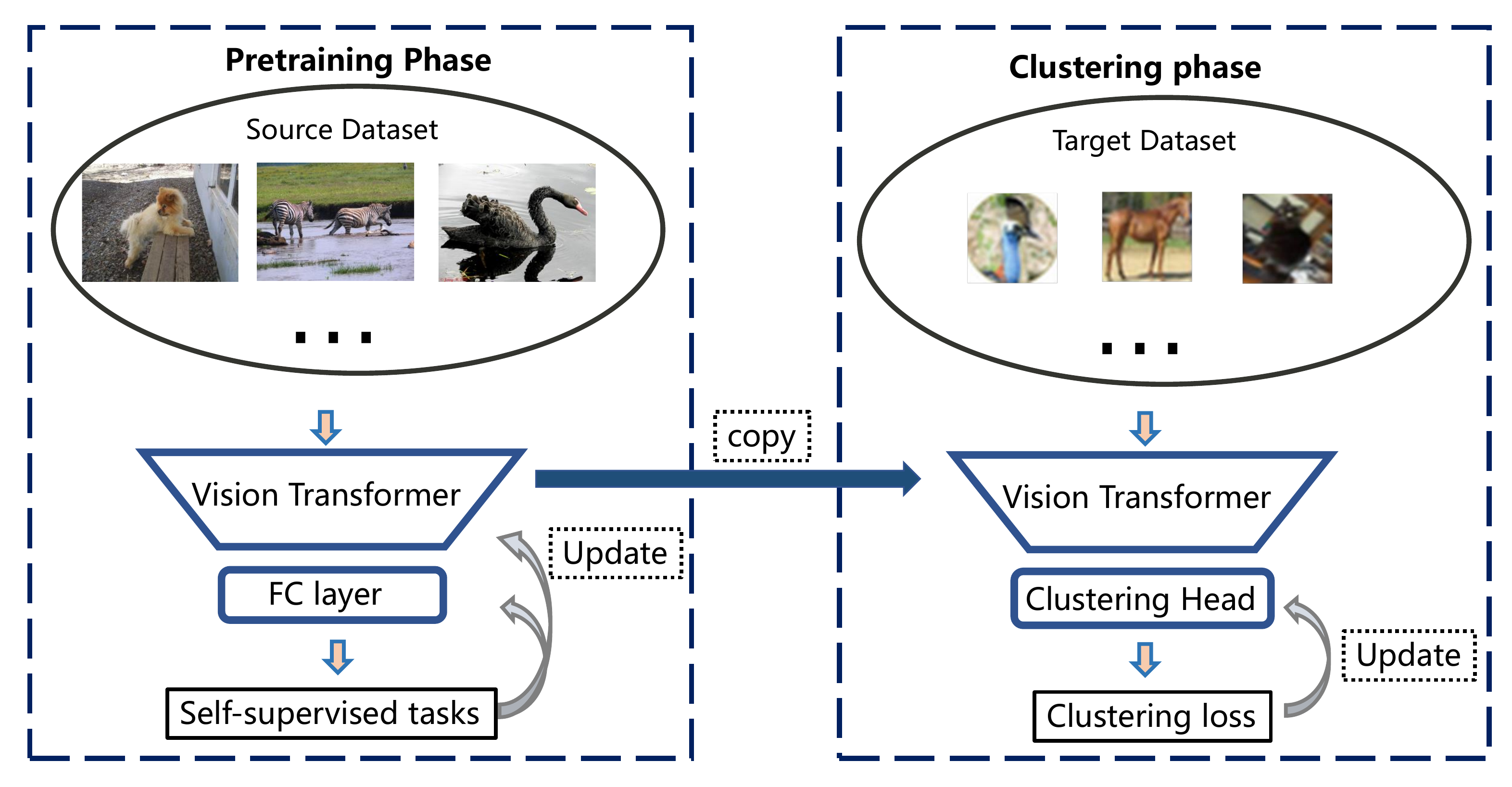}
	\caption{TSP: Two-stage image clustering with features from Self-supervised Pretraining.}
	\label{fig:process diagram}
\end{figure}

This paper is concerned with the task of clustering a  set $\mathcal{D}_{target}$ of images. We assume there is another unlabelled set  $\mathcal{D}_{source}$ of images that is potentially much larger. We propose to: (1) Train a feature extractor $f$ on $\mathcal{D}_{source}$ using self-supervised learning,
(2) map $\mathcal{D}_{target}$  to a latent space using $f$, and (3) perform clustering in the latent space.  This pipeline is called TSP and is illustrated in Figure
\ref{fig:process diagram}, where it is assumed that a ViT architecture is used for $f$.

Metaphorically, we think of the data points in $\mathcal{D}_{target}$ as black sheep, and semantic clusters among the data points as herds of black sheep.
We would like to place the black sheep onto an open grass field (the latent space) using a location assignment function $f$ such that the herd structures are maintained. We determine $f$ by considering how to place a much larger collection of white sheep $\mathcal{D}_{source}$ onto the same grass field. During the optimization process, $f$ changes from iteration to iteration and consequently the sheep would move around.  The key intuition is that, with the white sheep as cushions, the herd structures among the black sheep would likely be preserved.

In theory the feature mapping $f$ can be obtained using any self-supervised learning algorithm. We use DINO \cite{caron2021emerging} in this paper.  To perform clustering in the latent space, we use a loss function that is similar to SCAN \cite{van2020scan}.  Technical details of the two phases are given in the following two subsections.

\subsection{Feature Extractor Learning}
\label{sec. pretrain phase}

The feature extractor $f$ maps an input image $x$ into a latent vector
$f(x)$. We train it using DINO. In DINO, $f(x)$ is fed to another network (a shallow MLP)
called a projection head $h$  to get a second latent vector $g(x) = h(f(x))$.  A probability distribution $P(y|x)$ is then defined by applying softmax to $g(x)$ with a temperature parameter $\tau$. Here $y$ stands for a random variable whose possible values
are the  dimensions of the vector $g(x)$.
Let $\theta$ be the all the weights in $f$ and $g$.  Two copies of $\theta$ are maintained, $\theta_s$ for a student network and $\theta_t$ for a teacher network.

Data are augmented using the multi-crop strategy \cite{caron2020unsupervised}. From each input image $x$, a set $V$ of different views are generated. Two views $x_1^g$ and $x_2^g$ are global while the others are local.  At each iteration, the student weights ${\theta_s}$ are improved by taking a gradient step to minimize the sum of the following cross entropy loss over a minibatch of input images:

\begin{eqnarray}
 \sum_{v \in \{x^g_1, x^g_2\}} \sum_{v' \in V, v' \neq v}
-P_{\theta_t}(y|v) \log P_{\theta_s}(y|v').
\end{eqnarray}

\noindent The teacher is not trained. Its weights $\theta_t$ are determined as
an exponential moving average (EMA) of the student parameters:

\begin{eqnarray}
\theta_t \leftarrow \lambda \theta_t + (1-\lambda) \theta_s,
\end{eqnarray}

\noindent where $\lambda$ is picked according to a cosine schedule.  Consequently, the teacher is known as a momentum teacher. Note that the student is encouraged to match the teacher even when it has a local view of the input only. This discourages the model to pay attention to low-level details and helps with the extraction of semantically meaningful high-level features \cite{caron2020unsupervised}.

To avoid model collapse where both $P_{\theta_t}$ and $P_{\theta_s}$ become trivial distributions, a bias term $c$ is added $g_{\theta_t}(x)$. This operation is called centering and the center $c$ is updated for each minibatch $B$ as follows:

\begin{eqnarray}
c \leftarrow m c + (1-m) \frac{1}{|B|} \sum_{i: x_i \in B}g_{\theta_t}(x_i).
\end{eqnarray}

where $m>0$ is a rate parameter.  Centering prevents the distribution
$P_{\theta_t}$ from concentrating all the probability mass on one dimension.
To prevent $P_{\theta_t}$ from becoming the uniform distribution, sharpening is applied by lowering the temperature parameter $\tau_t$ for the teacher.

After DINO training, the final $f_{\theta_t}$ is used as the feature extractor. In our experiments, we use the feature extractors pretrained on ImageNet-1k (with the labels ignored) by the authors of DINO.

\subsection{Clustering in the Latent Space}
\label{sec.cluster-loss}
Assume the number of clusters is known and it is $N_c$.   Following
\cite{van2020scan, Dang_2021_CVPR}, we partition the data points in the latent space into $N_c$ clusters using a linear classifier $P_{\phi}(y|f(x))$ with parameters $\phi$.
Here $f(x)$  is the latent representation of input image $x$, and
$P_{\phi}(y|f(x))$  is a probability distribution over $N_c$ clusters.

It is not advisable to use a network with multiple layers for the
clustering head because it implies further feature transformation, and gives rise to another problem of cluster structure preservation, which is difficult to solve.
We also do not fine-tune the feature extractor on the target dataset (the black sheep) because it might negatively impact cluster structure preservation with the absence of the white sheep. Furthermore,
simply applying K-means on the latent space is  not ideal.
Empirical evidence in support of those arguments will be presented later.

In the following we introduce the loss function  for the classifier. For simplicity, we denote the latent representation $f(x)$ of an input image $x$  as $\tilde{x}$.
Let $\mathcal{N}_K(\tilde{x})$ be the $K$ nearest neighbors of $\tilde{x}$ in the latent space.
For each data point $\tilde{x}_i$, we define a probability distribution
$P(\tilde{x}_j|\tilde{x}_i)$ over the data points in the neighborhood of
$\tilde{x}_i$:

\begin{equation}
	\label{equation:W_ij}
	P(\tilde{x}_j|\tilde{x}_i) = \left\{ {\begin{array}{*{20}{l}}
			{\frac{1}{{{C_i}}}\exp \left( { - \frac{{\left\| {\tilde{x}_i - \tilde{x}_j } \right\|_2^2}}{{2\sigma_i^2}}} \right)}& \mbox{ if
				$\tilde{x}_j \in \mathcal{N}_K(\tilde{x}_i)$ }
			\\
			{0,}&{{\text{ otherwise }}}
	\end{array}} \right.
\end{equation}

where
$C_i$ is the normalization constant to ensure $\sum_j P(\tilde{x}_j|\tilde{x}_i) = 1$. The hyperparameter $\sigma_i$ controls the shape of the distribution. Following \cite{shaham2018spectralnet},  we  set it to be half of the distance between $\tilde{x}_i$ and the third nearest neighbor.  We optimize the classifier by minimizing the following loss:

\begin{eqnarray}
	\mathcal{L}_{cluster} = - \mathbf{E}_{\tilde{x}_i} \mathbf{E}_{\tilde{x}_j \sim P(\tilde{x}_j|\tilde{x}_i)} [
	\log \langle P_{\phi}(y|\tilde{x}_i),  P_{\phi}(y|\tilde{x}_j)   \rangle]
	- \lambda H\left( \mathbf{E}_{\tilde{x}_i} [P_{\phi}(y|\tilde{x}_i)] \right),
	\label{equation:clustering loss}
\end{eqnarray}

where $\langle, \rangle$ stands for inner product.
The first term encourages neighboring points be assigned to the same cluster. The second part is an   entropy regularization term. It helps to avoid trivial clustering where all data points are assigned to a single cluster.  Our loss function is
similar to the one used in SCAN \cite{van2020scan}, except that the distances between a data point $\tilde{x}_i$ and its neighbors are also taken into consideration.

The parameters $\phi$ of the clustering head can initialized either randomly or using K-means. In the latter case, the initial values of $\phi$ are calculated from the K-means cluster centers. The details are given in Appendix A.  Our empirical results indicate that K-means initialization is far superior.

\section{Experiments}

Our experiments are designed to show: 1).
TSP  can indeed leads to cluster-structure-preserving feature mappings  and  consequently superior clustering performances because of the adoption of the BaW strategy; 2)  The effectiveness of the BaW strategy depends on how robust the underlying architecture is to domain shift; 3). Clustering in the latent space of TSP requires a method  more sophisticated  than K-means; 4) Fine-tuning the   feature mapping of TSP might harm cluster structure preservation.

\subsection{Experimental Setups}

We compare three types of deep clustering methods:
1). E2E methods DEC \cite{xie2016unsupervised}, DAC \cite{chang2017deep}, DCMM \cite{wu2019deep}, IIC \cite{ji2019invariant}, and PICA \cite{huang2020deep}; 2). TSL methods
 SCAN \cite{van2020scan} and NNM \cite{Dang_2021_CVPR}; and 3). our TSP method.
Three instantiations of TSP are considered and they differ only in the
feature extractors used. The feature extractor are based on
the  ResNet50 \cite{he2016deep}, ViT-s 16 and ViT-b 16 \cite{touvron2021training} architectures, and were trained on ImageNet-1k (with the labels ignored) using DINO by the authors of DINO. The three versions of TSP will be referred to
as TSP-ResNet, TSP-ViT1 and TSP-ViT2 respectively. The hyperparameters for the clustering head
are chosen similarly to SCAN \cite{van2020scan}: The  $K$ in equation
(\ref{equation:W_ij}) is set at 20, and  the  $\lambda$ in
equation (\ref{equation:clustering loss}) is set at 3. It is trained using the Adam optimizer  with learning rate 0.0001, batch size  256, and total number of epochs  100. All the experiments were run on a GeForce RTX 2080 Ti machine. The training of a clustering head typically took less than an hour.

\subsection{Clustering Performances}

Following the common practice in the deep clustering literature,
we carry out the  evaluations on three (target) datasets CIFAR-10 \cite{krizhevsky2009learning}, CIFAR-100 \cite{krizhevsky2009learning} and STL-10 \cite{coates2011analysis}. The evaluation metrics include
 accuracy of cluster assignment (ACC),  normalized mutual information (NMI) and adjusted Rand index (ARI).  The results are shown in Table \ref{tabel:SOTA comparison}.  The row labeled
``supervised" shows results of supervised learning, which serve as upper bounds for clustering.
The last row will be discussed later.

% Table generated by Excel2LaTeX from sheet 'Result_recording'
\begin{table}[t]
	\centering
	\caption{Clustering performances of E2E, TSL, and TSP methods on three benchmarks. The standard deviations are computed from 10 independent runs.}	

	\begin{adjustbox}{width=1.0\columnwidth,center}
		\begin{tabular}{x{7em}x{3.15em}x{3.15em}x{3.15em}p{0.01em}x{3.15em}x{3.15em}x{3.15em}x{0.01em}x{3.15em}x{3.15em}x{3.15em}}
			\toprule
			\multicolumn{1}{c}{Datasets} & \multicolumn{3}{c}{CIFAR-10} &      & \multicolumn{3}{c}{CIFAR-100} &      & \multicolumn{3}{c}{STL-10} \\
			\cmidrule{2-4}\cmidrule{6-8}\cmidrule{10-12}    \multicolumn{1}{c}{Metrics} & NMI(\%)  & ACC(\%) & ARI(\%) &      & NMI(\%) & ACC(\%) & ARI(\%) &      & NMI(\%) &ACC(\%)& ARI(\%)\\ \ \\

			\multicolumn{1}{c}{K-Means } & 8.7  & 22.9 & 4.9  &      & 8.4  & 13   & 2.8  &      & 12.5 & 19.2 & 6.1 \\
			\multicolumn{1}{c}{DEC \cite{xie2016unsupervised} } & 25.7 & 30.1 & 16.1 &      & 13.6 & 18.5 & 5    &      & 27.6 & 35.9 & 18.6 \\
			\multicolumn{1}{c}{DAC \cite{chang2017deep}} & 39.6 & 52.2 & 30.6 &      & 18.5 & 23.8 & 8.8  &      & 36.6 & 47   & 25.7 \\
			\multicolumn{1}{c}{DCCM \cite{wu2019deep} } & 49.6 & 62.3 & 40.8 &      & 28.5 & 32.7 & 17.3 &      & 37.6 & 48.2 & 26.2 \\
			\multicolumn{1}{c}{IIC \cite{ji2019invariant}} & -    & 61.7 & -    &      & -    & 25.7 & -    &      & -    & 61   & - \\
			\multicolumn{1}{c}{PICA \cite{huang2020deep}} & 59.1 & 69.6 & 51.2 &      & 31   & 33.7 & 17.1 &      & 61.1 & 71.3 & 53.1 \\ \ \\

			\multicolumn{1}{c}{SCAN \cite{van2020scan}} & 71.5 & 81.6 & 66.5 &      & 44.9 & 44   & 28.3 &      & 67.3 & 79.2 & 61.8 \\
			\multicolumn{1}{c}{NNM \cite{Dang_2021_CVPR}} & 74.8 & 84.3 & 70.9 &      & 48.4 & 47.7 & 31.6 &      & 69.4 & 80.8 & 65 \\ \ \\

	\multicolumn{1}{c}{TSP-ResNet}
			& 50.8$\pm$1.1 & 58.5$\pm$2.0 & 40.2$\pm$1.1 &      & 40.3$\pm$1.3 & 39.0$\pm$1.8 & 22.8$\pm$1.5 &      & 74.5$\pm$2.4 & 78.9$\pm$4.8 & 66.8$\pm$4.2 \\

	 \rowcolor{pink!30!black!10}	TSP-ViT1  &84.7$\pm$1.0& 92.1$\pm$0.8&83.8$\pm$1.5&      & 58.2$\pm$1.4 & 54.9$\pm$2.5 & 40.8$\pm$2.1 &      & 94.1$\pm$1.4 & 97.0$\pm$1.2 & 93.8$\pm$2.3 \\
			
	 \rowcolor{pink!30!black!10}	TSP-ViT2 & \textbf{88.0}$\pm$1.2 & \textbf{94.0}$\pm$1.3 & \textbf{87.5}$\pm$2.3 &      &\textbf{61.4}$\pm$1.4 & \textbf{55.6}$\pm$2.5 & \textbf{43.3}$\pm$1.8 &      & \textbf{95.8}$\pm$2.0 & \textbf{97.9}$\pm$1.7 & \textbf{95.6}$\pm$3.1 \\ \ \\

 \rowcolor{SeaGreen3!30!} \multicolumn{1}{c}{Supervised}   & 91.5 & 96.6 &92.2 &      & 81.9 & 89.2 & 78.9 &      & 97.0 & 98.9 & 97.6 \\

TSP K-means &  77.5$\pm$2.2 & 74.9$\pm$6.2 & 67.7$\pm$4.6 &      & 56.2$\pm$1.2 & 50.9$\pm$2.8 & 35.7$\pm$1.9 &      & 89.2$\pm$3.0 & 81.7$\pm$9.2 & 80.2$\pm$8.1 \\

			\bottomrule
		\end{tabular}%
		\label{tab:addlabel}%
	\end{adjustbox}	
			\label{tabel:SOTA comparison}
	
\end{table}%

The TSL methods SCAN and NNM represent the previous state-of-the-art in deep clustering. We see that TSP-ViT1 and TSP-ViT2 outperform SCAN and NMM by large margins on all three datasets and under all evaluation metrics.
On the other hand,  the performances of TSP-ResNet are mixed. It is inferior to SCAN and NNM on CIFAR 10 and CIFAR 100, although it is slightly better on STL-10.
We will examine the reasons for the performance differences in the next few subsections.
To start with, the differences between TSP-ViT1 beats TSP-ViT2 is evidently because  ViT-b 16 is larger than  ViT-s 16 although they have the same type of architecture.

\subsection{Cluster Structure Preservation}
\label{sec: cluster structure preservation}

Why do TSP-ViT1 and TSP-ViT2 perform so well?
We believe a key reason is that their feature mappings preserve cluster structures  well.
We have seen some evidence for the belief already in Figures
\ref{fig:consistency of nearest neighbors} and \ref{fig:nearest_neighbors_visualization}. With high probabilities,  image pairs placed close to each other in the latent space by the feature mappings are from the same semantic cluster in the image space.  This is true for all the three datasets.

\begin{figure}[t]
	\centering

	\subfigure[Pretrained on ImageNet-1k ]{	\includegraphics[height=0.415 \linewidth]{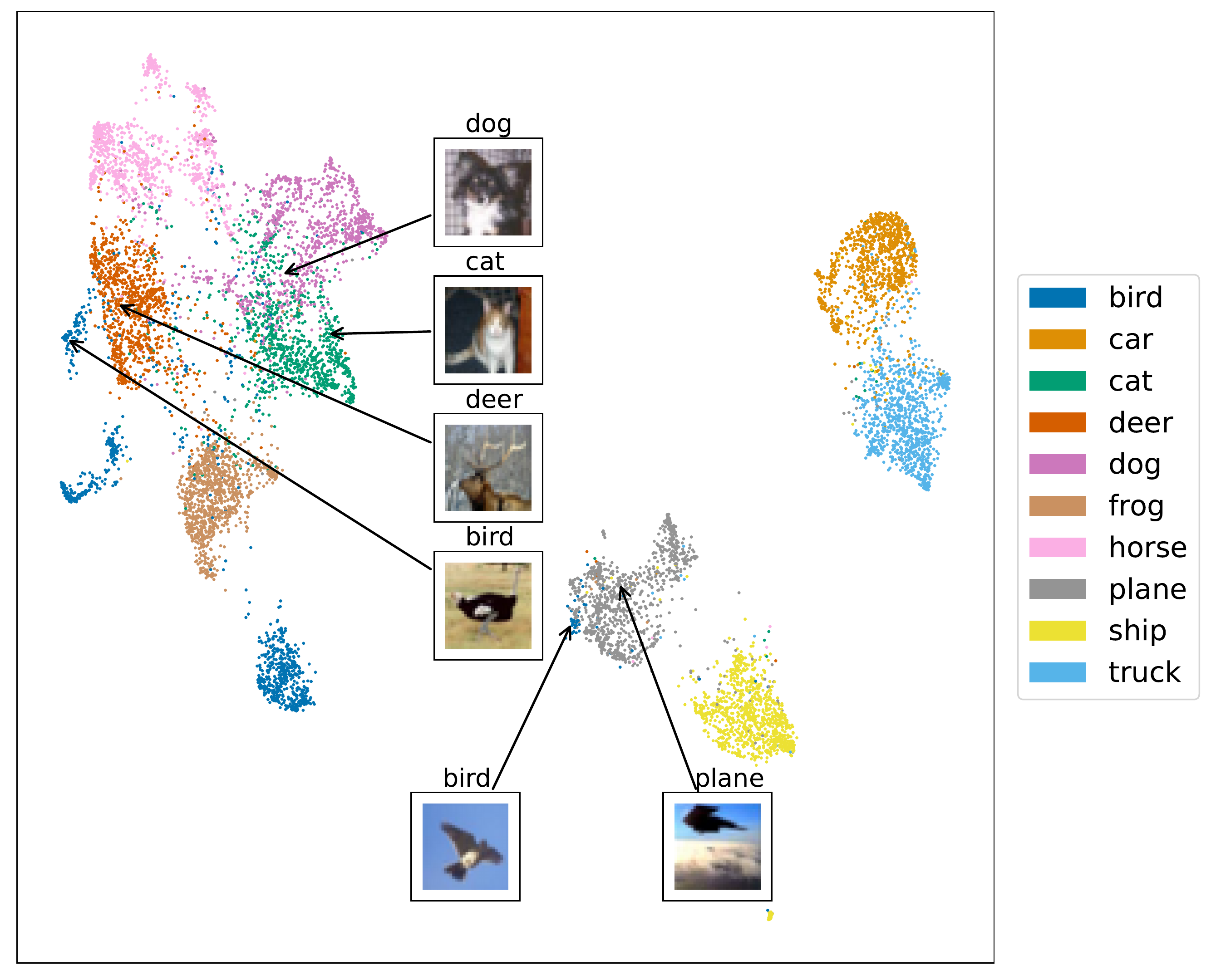}
		\label{fig:umap visualization ViTs pretrained on ImageNet-1k}
	}
\hfill
	\subfigure[Pretrained on CIFAR-10 ]{
	\includegraphics[height=0.415\linewidth]{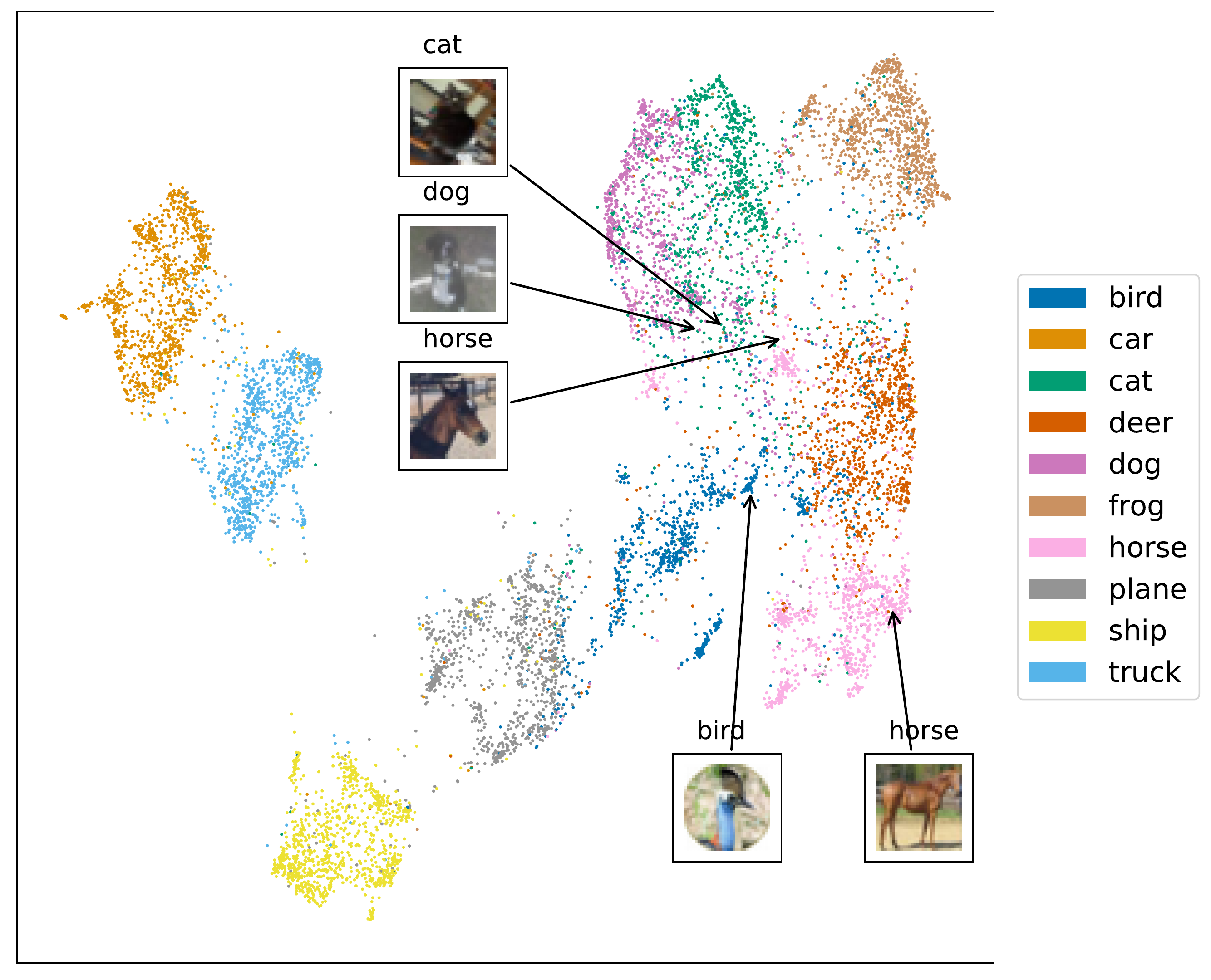}
	\label{fig:umap visualization ViTs pretrained on CIFAR-10}
}
	\caption{Distribution of CIFAR-10 data points  in the latent space of TSP-ViT1. A ViT feature mapping trained on ImageNet-1k preserves the cluster structure of CIFAR-10 than another trained on CIFAR-10 itself. }
	\label{figure:umap visualization ViTs pretrained on different source datasets}
\end{figure}

How do the data points distribute in the latent space?
Figure
\ref{figure:umap visualization ViTs pretrained on different source datasets} (a) shows a UMAP visualization \cite{mcinnes2018umap} of
 the CIFAR-10 data points in the latent space of TSP-ViT1.
We see that the 10 semantic clusters are mostly well separated.
There are imperfections. For example,
the cat and dog clusters are not clearly separated.  Some bird images (flying) are placed into the plane cluster, while other bird images (austrich)  are placed close to the deer cluster.  All those seem reasonable visually.

The extent to which a feature mapping preserves cluster structures depends
heavily on the pretraining data  used. Figure \ref{figure:umap visualization ViTs pretrained on different source datasets} (a) is
for a ViT feature mapping pretrained on ImageNet-1k. If it was pretrained on CIFAR-10 instead,   data distribution in the latent space would change significantly, as shown in Figure \ref{figure:umap visualization ViTs pretrained on different source datasets} (b).
   It is clear that in this case the semantic clusters are not as well separated as in (a). Six semantic clusters form a nebula on the right.
   We trained a clustering head  on this latent space.    The ACC is 80.2\%, which is a drastic  drop from  92.1\% of the previous case.

In a TSP deep clustering method, the feature mapping is pretrained on
one dataset (the white sheep) and is applied to another dataset (the black ship).  Domain shift is clearly an issue, and it is hence important to use
an architecture that is relatively robust to domain shift. Our experiments indicate that the ViT architecture seems to be a good choice. In fact,
TSP-ViT1 was pretrained on ImageNet-1k and it works well CIFAR-10.  This is consistent with previous observations in  supervised learning that ViT is relatively robust to domain shift \cite{paul2021vision}.  

\begin{figure}[t]
	\centering

	\subfigure[Target dataset STL-10]{	\includegraphics[width=0.45 \linewidth]{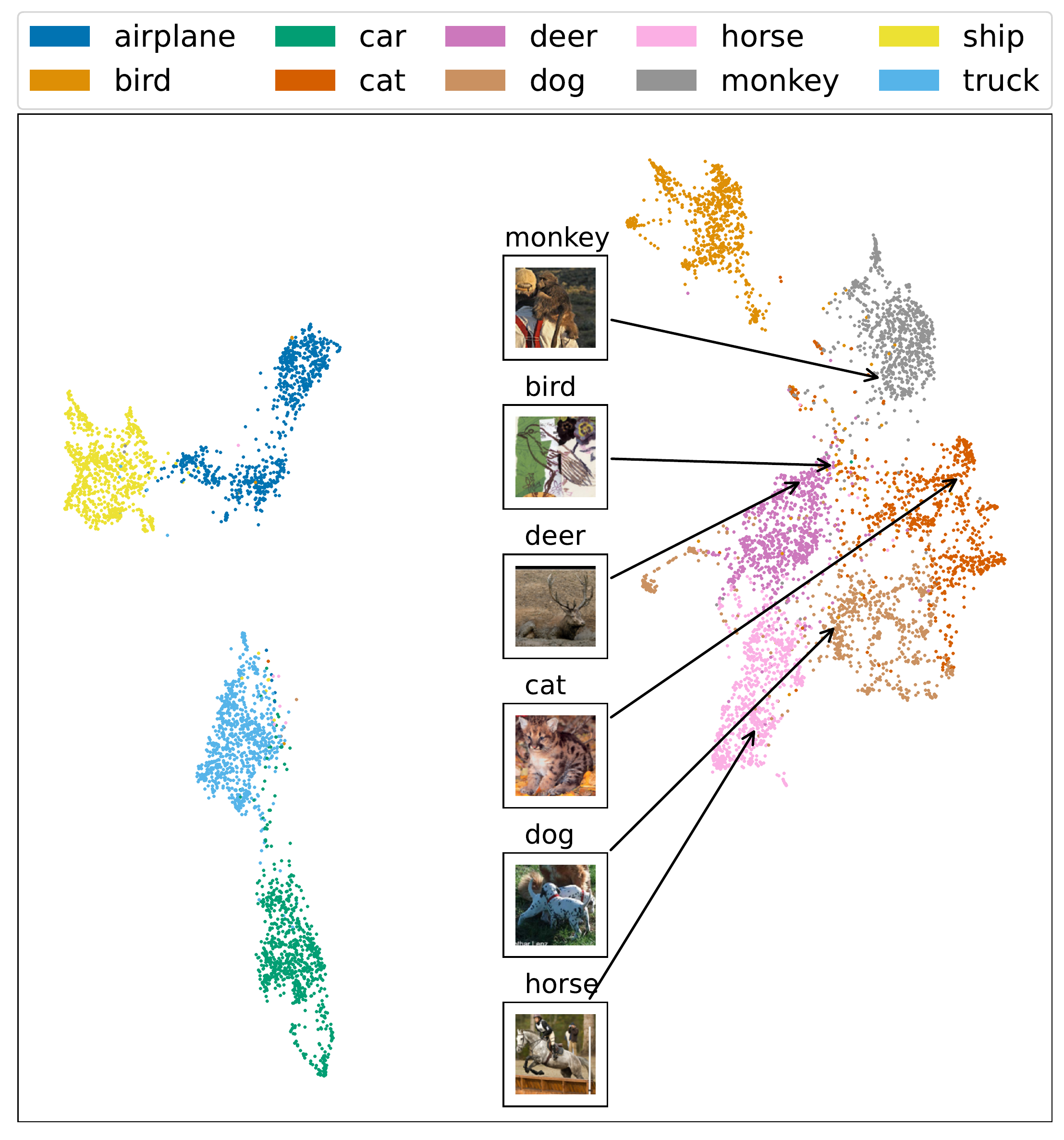}
		\label{fig:umap visualization ResNet50 STL10}
	}
\hfill
	\subfigure[Target dataset CIFAR-10]{
	\includegraphics[width=0.45\linewidth]{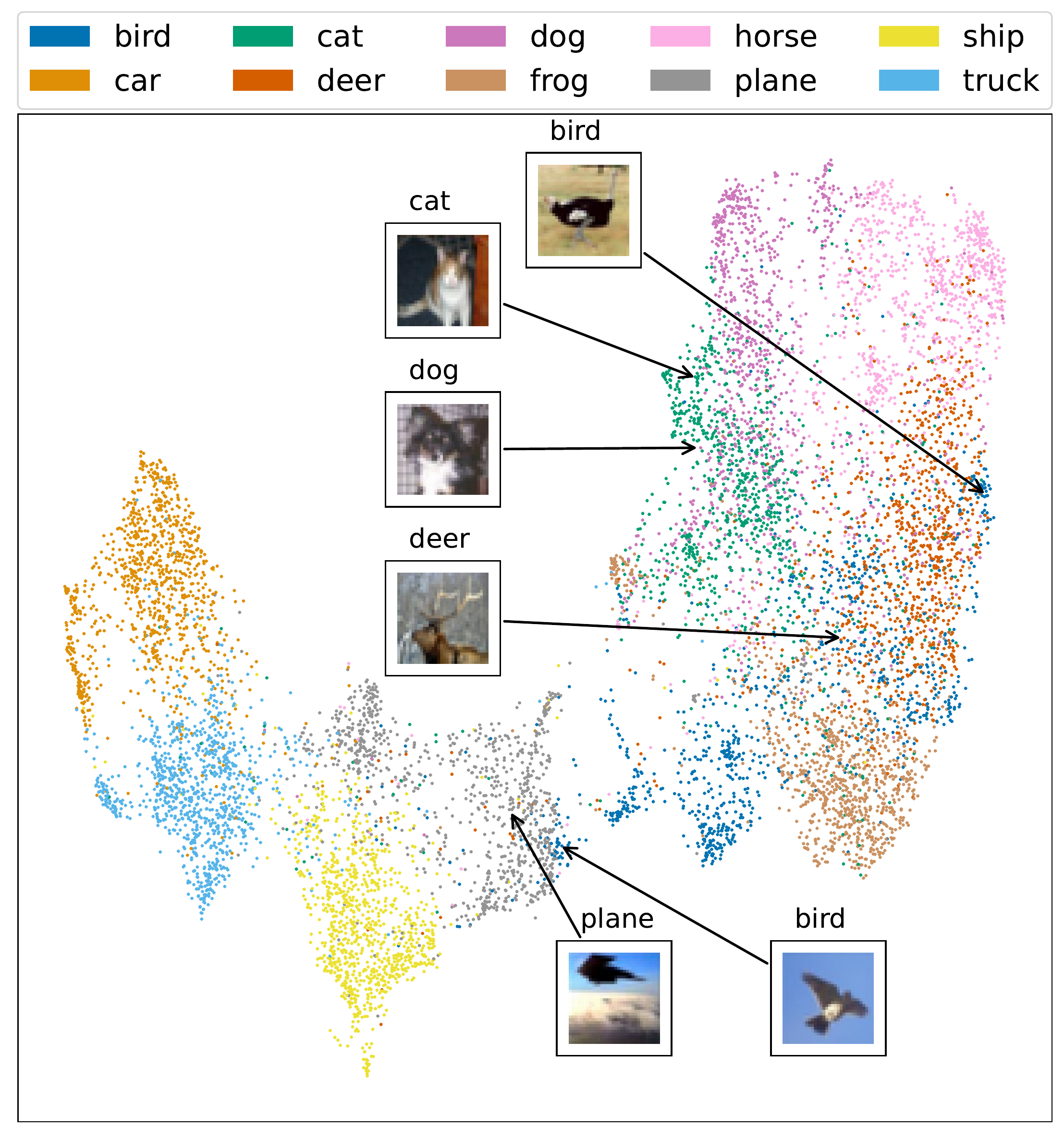}
	\label{fig:umap visualization ResNet 50 + CIFAR-10}
}
	\caption{Distribution of the STL-10 and CIFAR-10 data points in the
latent space of TSP-ResNet. A ResNet feature map trained on ImageNet-1k preserves the cluster structures of STL-10, a subset of ImageNet-1,  better than those of CIFAR-10.  }
	\label{figure:umap visualization ResNet50}
\end{figure}

 The situation with  TSP-ResNet is different. The feature mapping of  TSP-ResNet was also pretrained on ImageNet-1k. It preserves the cluster structures of STL-10 relatively well
 as shown in Figure \ref{figure:umap visualization ResNet50} (a), probably because STL-10 is a subset of ImageNet-1k.
 However, it does a much worse job in preserving the cluster structures of CIFAR-10   as shown in Figure \ref{figure:umap visualization ResNet50} (b).  This explains its inferior
 clustering performance on CIFAR-10.  It has been observed
 in  supervised learning that ResNet is not robust to
domain shift \cite{recht2019imagenet}.

\subsection{Clustering Head and Fine-tuning}

The simplest way to cluster data points in the latent space is to run K-means.  As shown in the last row of Table
\ref{tabel:SOTA comparison}, this can lead to competitive clustering results
when a strong feature mapping is used.
However, we can usually achieve better results using a clustering method more sophisticated than K-means as it clear from Table
 \ref{tabel:SOTA comparison}.  The reason is that the natural clusters in the latent space are not globular as shown in Figures
\ref{figure:umap visualization ViTs pretrained on different source datasets} (a) and
\ref{figure:umap visualization ResNet50} (a).

In supervised learning, a pretrained model usually needs
to be fine-tuned to achieved good performance on downstream tasks \cite{devlin2018bert}. Contrary to this common wisdom, we found that fine-tuning the feature mapping of TSP is counter-productive.  In fact, the ACC of TSP-ViT1 on CIFAR-10 drops drastically to 65.7\% from 92.1\% when its feature extractor is fine-tuned
during the clustering phase.

\begin{figure}[t]
	\centering
\includegraphics[height=0.415 \linewidth]{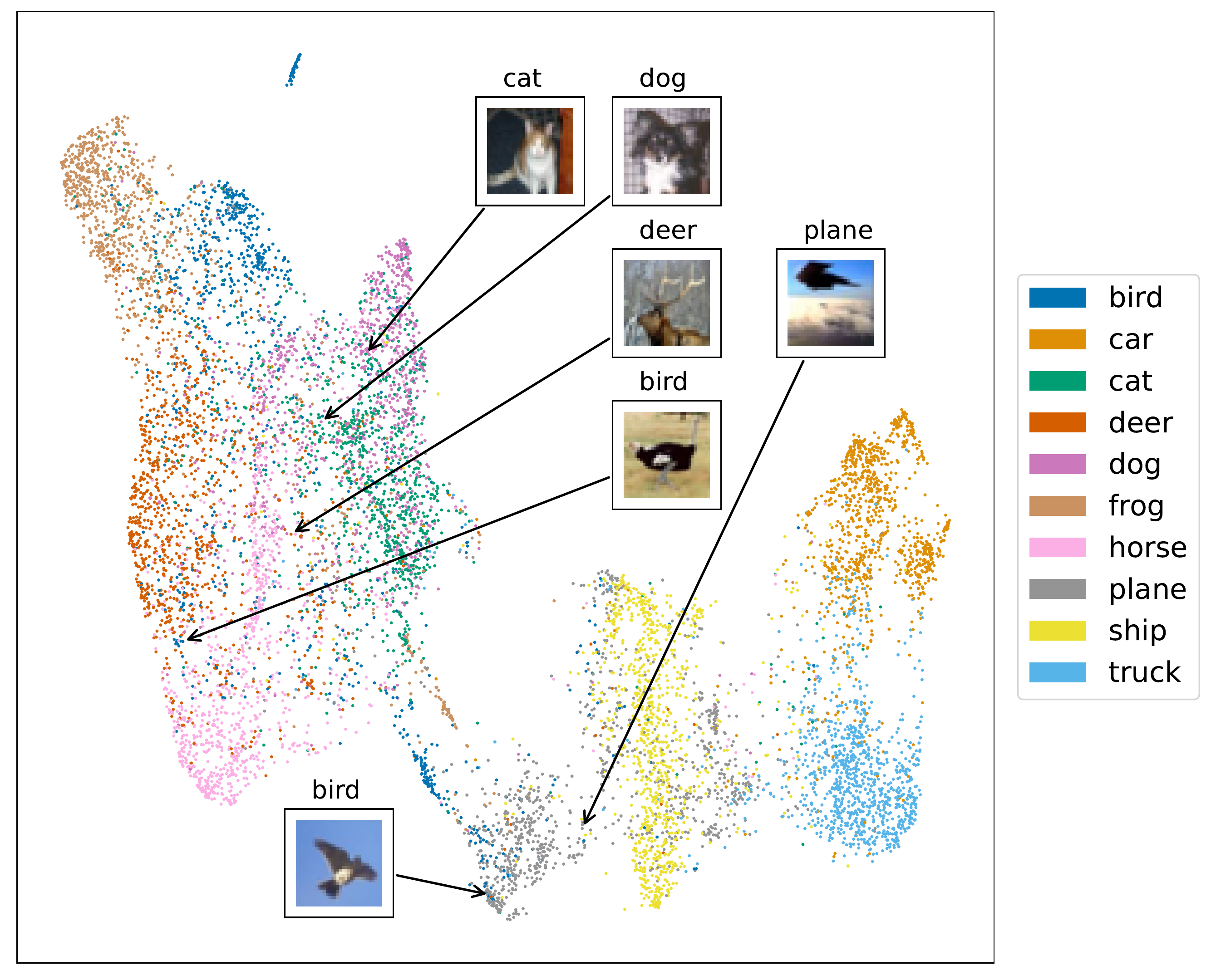}

	\caption{Distribution of CIFAR-10 data points in the latent space of TSP-ViT1 after feature extractor fine-tuning. The  cluster structures in \ref{figure:umap visualization ViTs pretrained on different source datasets} (a) are severely damaged }
	\label{figure6}
\end{figure}

To explain the phenomenon, we refer to the BaW metaphor again.  Suppose the herds
of black sheep are originally well separated in the latent space.
During fine-tuning, the feature mapping changes from iteration to iteration, and hence the black sheep move around.   With no white sheep as cushions, different herds of black sheep can easily merge, destroying the initial cluster structures.  Figure \ref{figure6} shows the distribution of CIFAR-10 data points
in the latent space after the feature mapping is fine-tuned.  We see that
the  cluster structures in \ref{figure:umap visualization ViTs pretrained on different source datasets} (a) are severely damaged.

\subsection{Other Results}

Although K-means is too simplistic for clustering in the latent space, it provides a good way to initialize the clustering head of TSP.  In fast, the ACC of TSP-ViT1 on CIFAR-10 drops to 85.6\% from 92.1\% when the clustering head is initialized randomly instead of using K-means.

In all the experiments reported above, the number $K$ of nearest neighbors
is set at 20.  It turns out that  TSP-ViT1 and TSP-ViT2 are not sensitive to $K$.  Table \ref{table:effect number of negihbors} show the results of TSP-ViT1 on CIFAR-10 for various choices of $K$.

	\begin{table}[h!]
		\centering

		\caption{Performances of TSP-ViT1 on CIFAR-10 for various choices of $K$.}
		
			%\begin{adjustbox}{width=0.8\columnwidth,center}
		\begin{tabular}{ccll}
			\toprule
			 \# of neighbors & ACC(\%)  & \multicolumn{1}{c}{NMI(\%)} & \multicolumn{1}{c}{ARI(\%)} \\
			\midrule
			 5    & \multicolumn{1}{l}{92.41$\pm$1.01} & 85.10$\pm$1.25 & 84.33$\pm$1.85 \\
			 10   & \multicolumn{1}{l}{92.59$\pm$0.87} & 85.31$\pm$1.08 & 84.64$\pm$1.57 \\
			 15   & \multicolumn{1}{l}{92.25$\pm$0.98} & 84.84$\pm$1.16 & 84.01$\pm$1.73 \\
			 20   & \multicolumn{1}{l}{92.11$\pm$0.88} & 84.59$\pm$1.04 & 83.72$\pm$1.55 \\
			 25   & 92.03$\pm$0.91 & 84.50$\pm$1.10 & 83.58$\pm$1.62 \\
			 30   & 91.88$\pm$0.90 & 84.23$\pm$1.12 & 83.27$\pm$1.59 \\
			\bottomrule
		\end{tabular}%
	%\end{adjustbox}
	
		\label{table:effect number of negihbors}
	\end{table}%

	\section{Conclusion}

In deep clustering, we first map data points to a latent space using a neural network and then group the data points into cluster there.  It is of critical importance that cluster structures are preserved in the first step. However, cluster structure preservation is difficult to achieve due to the flexibility of neural networks.  In this paper, we propose black-sheep-among-white-sheep (BaW) as a general strategy to learning cluster-structure-preserving feature mappings. The idea is to pretrain a feature extractor on a potentially very large dataset (the white sheep) via self-supervised learning, use it to map a target dataset (the black sheep) to a latent space, and perform clustering there.
When instantiated with the ViT architecture and  the self-supervised learning algorithm DINO,  our TSP method  has achieved results drastically better than the previous state-of-the art. We have also gained some interesting insights. One insight is that TSP requires an architecture that is relatively robust to domain shift.  For example, TSP does not work well with ResNet.
Another insight is that, contrary to the case of supervised learning, fine-tuning the feature mapping is counter-productive in deep clustering.

Similar to previous work on deep clustering, the benchmark datasets used in our work are balanced in the sense that the sizes of different ground-truth clusters are the same.  An interesting direction for future research is to develop methods that work well on imbalanced datasets. We believe the BaW strategy can contribute to the solution of the problem. The presence of white sheep should help us to identify black sheep clusters of different sizes.

\bibliography{biblio}

\begin{thebibliography}{34}
\providecommand{\natexlab}[1]{#1}
\providecommand{\url}[1]{\texttt{#1}}
\expandafter\ifx\csname urlstyle\endcsname\relax
  \providecommand{\doi}[1]{doi: #1}\else
  \providecommand{\doi}{doi: \begingroup \urlstyle{rm}\Url}\fi

\bibitem[He et~al.(2015)He, Zhang, Ren, and Sun]{he2015delving}
Kaiming He, Xiangyu Zhang, Shaoqing Ren, and Jian Sun.
\newblock Delving deep into rectifiers: Surpassing human-level performance on
  imagenet classification.
\newblock In \emph{Proceedings of the IEEE international conference on computer
  vision}, pages 1026--1034, 2015.

\bibitem[Dilokthanakul et~al.(2016)Dilokthanakul, Mediano, Garnelo, Lee,
  Salimbeni, Arulkumaran, and Shanahan]{dilokthanakul2016deep}
Nat Dilokthanakul, Pedro~AM Mediano, Marta Garnelo, Matthew~CH Lee, Hugh
  Salimbeni, Kai Arulkumaran, and Murray Shanahan.
\newblock Deep unsupervised clustering with gaussian mixture variational
  autoencoders.
\newblock \emph{arXiv preprint arXiv:1611.02648}, 2016.

\bibitem[Guo et~al.(2017)Guo, Gao, Liu, and Yin]{guo2017improved}
Xifeng Guo, Long Gao, Xinwang Liu, and Jianping Yin.
\newblock Improved deep embedded clustering with local structure preservation.
\newblock In \emph{International Joint Conference on Artificial Intelligence},
  pages 1753--1759, 2017.

\bibitem[Jiang et~al.(2017)Jiang, Zheng, Tan, Tang, and
  Zhou]{jiang2017variational}
Zhuxi Jiang, Yin Zheng, Huachun Tan, Bangsheng Tang, and Hanning Zhou.
\newblock Variational deep embedding: an unsupervised and generative approach
  to clustering.
\newblock In \emph{International Joint Conference on Artificial Intelligence},
  pages 1965--1972, 2017.

\bibitem[Xie et~al.(2016)Xie, Girshick, and Farhadi]{xie2016unsupervised}
Junyuan Xie, Ross Girshick, and Ali Farhadi.
\newblock Unsupervised deep embedding for clustering analysis.
\newblock In \emph{International Conference on Machine Learning}, pages
  478--487. PMLR, 2016.

\bibitem[Yang et~al.(2017)Yang, Fu, Sidiropoulos, and Hong]{yang2017towards}
Bo~Yang, Xiao Fu, Nicholas~D Sidiropoulos, and Mingyi Hong.
\newblock Towards k-means-friendly spaces: Simultaneous deep learning and
  clustering.
\newblock In \emph{international conference on machine learning}, pages
  3861--3870, 2017.

\bibitem[Huang et~al.(2014)Huang, Huang, Wang, and Wang]{huang2014deep}
Peihao Huang, Yan Huang, Wei Wang, and Liang Wang.
\newblock Deep embedding network for clustering.
\newblock In \emph{International conference on pattern recognition}, pages
  1532--1537. IEEE, 2014.

\bibitem[Chen et~al.(2017)Chen, Lv, and Zhang]{chen2017unsupervised}
Dongdong Chen, Jiancheng Lv, and Yi~Zhang.
\newblock Unsupervised multi-manifold clustering by learning deep
  representation.
\newblock In \emph{Workshops at the thirty-first AAAI conference on artificial
  intelligence}, 2017.

\bibitem[Van~Gansbeke et~al.(2020)Van~Gansbeke, Vandenhende, Georgoulis,
  Proesmans, and Van~Gool]{van2020scan}
Wouter Van~Gansbeke, Simon Vandenhende, Stamatios Georgoulis, Marc Proesmans,
  and Luc Van~Gool.
\newblock Scan: Learning to classify images without labels.
\newblock In \emph{European Conference on Computer Vision}, pages 268--285,
  2020.

\bibitem[Dang et~al.(2021)Dang, Deng, Yang, Wei, and Huang]{Dang_2021_CVPR}
Zhiyuan Dang, Cheng Deng, Xu~Yang, Kun Wei, and Heng Huang.
\newblock Nearest neighbor matching for deep clustering.
\newblock In \emph{Proceedings of the IEEE/CVF Conference on Computer Vision
  and Pattern Recognition}, pages 13693--13702, 2021.

\bibitem[Han et~al.(2020)Han, Park, Park, Kim, and
  Cha]{hanMitigatingEmbeddingClass2020}
Sungwon Han, Sungwon Park, Sungkyu Park, Sundong Kim, and Meeyoung Cha.
\newblock Mitigating embedding and class assignment mismatch in unsupervised
  image classification.
\newblock In \emph{European Conference on Computer Vision}, pages 768--784,
  2020.

\bibitem[Devlin et~al.(2019)Devlin, Chang, Lee, and Toutanova]{devlin2018bert}
Jacob Devlin, Ming-Wei Chang, Kenton Lee, and Kristina Toutanova.
\newblock {BERT}: Pre-training of deep bidirectional transformers for language
  understanding.
\newblock In \emph{North American Chapter of the Association for Computational
  Linguistics}, pages 4171--4186, 2019.

\bibitem[He et~al.(2019)He, Girshick, and Doll{\'a}r]{he2019rethinking}
Kaiming He, Ross Girshick, and Piotr Doll{\'a}r.
\newblock Rethinking imagenet pre-training.
\newblock In \emph{Proceedings of the IEEE/CVF International Conference on
  Computer Vision}, pages 4918--4927, 2019.

\bibitem[He et~al.(2016)He, Zhang, Ren, and Sun]{he2016deep}
Kaiming He, Xiangyu Zhang, Shaoqing Ren, and Jian Sun.
\newblock Deep residual learning for image recognition.
\newblock In \emph{Proceedings of the IEEE conference on computer vision and
  pattern recognition}, pages 770--778, 2016.

\bibitem[Touvron et~al.(2021)Touvron, Cord, Douze, Massa, Sablayrolles, and
  J{\'e}gou]{touvron2021training}
Hugo Touvron, Matthieu Cord, Matthijs Douze, Francisco Massa, Alexandre
  Sablayrolles, and Herv{\'e} J{\'e}gou.
\newblock Training data-efficient image transformers \& distillation through
  attention.
\newblock In \emph{International Conference on Machine Learning}, pages
  10347--10357, 2021.

\bibitem[Russakovsky et~al.(2015)Russakovsky, Deng, Su, Krause, Satheesh, Ma,
  Huang, Karpathy, Khosla, Bernstein, et~al.]{russakovsky2015imagenet}
Olga Russakovsky, Jia Deng, Hao Su, Jonathan Krause, Sanjeev Satheesh, Sean Ma,
  Zhiheng Huang, Andrej Karpathy, Aditya Khosla, Michael Bernstein, et~al.
\newblock Imagenet large scale visual recognition challenge.
\newblock \emph{International journal of computer vision}, 115\penalty0
  (3):\penalty0 211--252, 2015.

\bibitem[Caron et~al.(2021)Caron, Touvron, Misra, J{\'e}gou, Mairal,
  Bojanowski, and Joulin]{caron2021emerging}
Mathilde Caron, Hugo Touvron, Ishan Misra, Herv{\'e} J{\'e}gou, Julien Mairal,
  Piotr Bojanowski, and Armand Joulin.
\newblock Emerging properties in self-supervised vision transformers.
\newblock In \emph{Proceedings of the IEEE/CVF International Conference on
  Computer Vision}, pages 9650--9660, 2021.

\bibitem[Naseer et~al.(2021)Naseer, Ranasinghe, Khan, Hayat, Shahbaz~Khan, and
  Yang]{naseer2021intriguing}
Muhammad~Muzammal Naseer, Kanchana Ranasinghe, Salman~H Khan, Munawar Hayat,
  Fahad Shahbaz~Khan, and Ming-Hsuan Yang.
\newblock Intriguing properties of vision transformers.
\newblock \emph{Advances in Neural Information Processing Systems}, 34, 2021.

\bibitem[Chang et~al.(2017)Chang, Wang, Meng, Xiang, and Pan]{chang2017deep}
Jianlong Chang, Lingfeng Wang, Gaofeng Meng, Shiming Xiang, and Chunhong Pan.
\newblock Deep adaptive image clustering.
\newblock In \emph{Proceedings of the IEEE International Conference on Computer
  Vision}, pages 5879--5887, 2017.

\bibitem[Wu et~al.(2019)Wu, Long, Wang, Qian, Li, Lin, and Zha]{wu2019deep}
Jianlong Wu, Keyu Long, Fei Wang, Chen Qian, Cheng Li, Zhouchen Lin, and
  Hongbin Zha.
\newblock Deep comprehensive correlation mining for image clustering.
\newblock In \emph{Proceedings of the IEEE/CVF International Conference on
  Computer Vision}, pages 8150--8159, 2019.

\bibitem[Ji et~al.(2019)Ji, Henriques, and Vedaldi]{ji2019invariant}
Xu~Ji, Joao~F Henriques, and Andrea Vedaldi.
\newblock Invariant information clustering for unsupervised image
  classification and segmentation.
\newblock In \emph{Proceedings of the IEEE/CVF International Conference on
  Computer Vision}, pages 9865--9874, 2019.

\bibitem[Huang et~al.(2020)Huang, Gong, and Zhu]{huang2020deep}
Jiabo Huang, Shaogang Gong, and Xiatian Zhu.
\newblock Deep semantic clustering by partition confidence maximisation.
\newblock In \emph{Proceedings of the IEEE/CVF Conference on Computer Vision
  and Pattern Recognition}, pages 8849--8858, 2020.

\bibitem[Gidaris et~al.(2018)Gidaris, Singh, and
  Komodakis]{gidaris2018unsupervised}
Spyros Gidaris, Praveer Singh, and Nikos Komodakis.
\newblock Unsupervised representation learning by predicting image rotations.
\newblock In \emph{International Conference on Learning Representations}, 2018.

\bibitem[Noroozi and Favaro(2016)]{noroozi2016unsupervised}
Mehdi Noroozi and Paolo Favaro.
\newblock Unsupervised learning of visual representations by solving jigsaw
  puzzles.
\newblock In \emph{European Conference on Computer Vision}, pages 69--84, 2016.

\bibitem[Caron et~al.(2018)Caron, Bojanowski, Joulin, and Douze]{caron2018deep}
Mathilde Caron, Piotr Bojanowski, Armand Joulin, and Matthijs Douze.
\newblock Deep clustering for unsupervised learning of visual features.
\newblock In \emph{European Conference on Computer Vision}, pages 132--149,
  2018.

\bibitem[Chen et~al.(2020)Chen, Kornblith, Swersky, Norouzi, and
  Hinton]{chen2020big}
Ting Chen, Simon Kornblith, Kevin Swersky, Mohammad Norouzi, and Geoffrey~E
  Hinton.
\newblock Big self-supervised models are strong semi-supervised learners.
\newblock \emph{Advances in Neural Information Processing Systems},
  33:\penalty0 22243--22255, 2020.

\bibitem[Caron et~al.(2020)Caron, Misra, Mairal, Goyal, Bojanowski, and
  Joulin]{caron2020unsupervised}
Mathilde Caron, Ishan Misra, Julien Mairal, Priya Goyal, Piotr Bojanowski, and
  Armand Joulin.
\newblock Unsupervised learning of visual features by contrasting cluster
  assignments.
\newblock \emph{Advances in Neural Information Processing Systems},
  33:\penalty0 9912--9924, 2020.

\bibitem[Grill et~al.(2020)Grill, Strub, Altch{\'e}, Tallec, Richemond,
  Buchatskaya, Doersch, Avila~Pires, Guo, Gheshlaghi~Azar,
  et~al.]{grill2020bootstrap}
Jean-Bastien Grill, Florian Strub, Florent Altch{\'e}, Corentin Tallec, Pierre
  Richemond, Elena Buchatskaya, Carl Doersch, Bernardo Avila~Pires, Zhaohan
  Guo, Mohammad Gheshlaghi~Azar, et~al.
\newblock Bootstrap your own latent-a new approach to self-supervised learning.
\newblock \emph{Advances in Neural Information Processing Systems},
  33:\penalty0 21271--21284, 2020.

\bibitem[Shaham et~al.(2018)Shaham, Stanton, Li, Basri, Nadler, and
  Kluger]{shaham2018spectralnet}
Uri Shaham, Kelly Stanton, Henry Li, Ronen Basri, Boaz Nadler, and Yuval
  Kluger.
\newblock Spectralnet: Spectral clustering using deep neural networks.
\newblock In \emph{International Conference on Learning Representations}, 2018.

\bibitem[Krizhevsky et~al.(2009)]{krizhevsky2009learning}
Alex Krizhevsky et~al.
\newblock Learning multiple layers of features from tiny images.
\newblock 2009.

\bibitem[Coates et~al.(2011)Coates, Ng, and Lee]{coates2011analysis}
Adam Coates, Andrew Ng, and Honglak Lee.
\newblock An analysis of single-layer networks in unsupervised feature
  learning.
\newblock In \emph{Artificial Intelligence and Statistics}, pages 215--223,
  2011.

\bibitem[McInnes et~al.(2018)McInnes, Healy, and Melville]{mcinnes2018umap}
Leland McInnes, John Healy, and James Melville.
\newblock Umap: Uniform manifold approximation and projection for dimension
  reduction.
\newblock \emph{arXiv preprint arXiv:1802.03426}, 2018.

\bibitem[Paul and Chen(2022)]{paul2021vision}
Sayak Paul and Pin-Yu Chen.
\newblock Vision transformers are robust learners.
\newblock In \emph{Association for the Advancement of Artificial Intelligence},
  2022.

\bibitem[Recht et~al.(2019)Recht, Roelofs, Schmidt, and
  Shankar]{recht2019imagenet}
Benjamin Recht, Rebecca Roelofs, Ludwig Schmidt, and Vaishaal Shankar.
\newblock Do imagenet classifiers generalize to imagenet?
\newblock In \emph{International Conference on Machine Learning}, pages
  5389--5400, 2019.

\end{thebibliography}

\clearpage
\appendix
\section{Initialization of Clustering Head}
Initialization of clustering head decides the clustering performance, since bad initialization of clustering head introduces incorrect signal on the beginning stage and leads to sub-optimal results. To get a better and stable clustering performance we adopt K-means to initialize the weights of clustering head. 

The centers of K-means on the latent space are used for the weights of clustering head with a scaling fact. 1) We apply K-means on latent space of datas $\{f(x)\}$ to get the centers $\{\mathcal{C}_i| i=1,2,...N_c\}$, and $\mathcal{C}_i$ is a vector with the same dimension as the $f(x)$, denoting this dimension as $D$. 2) Initialize the weights of clustering head $\phi \in \mathbb{R}^{N_c\times D}$ by $\phi_i = \frac{\mathcal{C}_i}{||\mathcal{C}_i|| \sqrt{D}}, i=1,2,...N_c$, where $\phi_i$ is $i-th$ row of $\phi$.  

\section{Case study on Clustering Results}
\begin{figure}[H]
	\centering

	\subfigure[Source dataset: CIFAR10]{
		\includegraphics[height=0.4\linewidth]{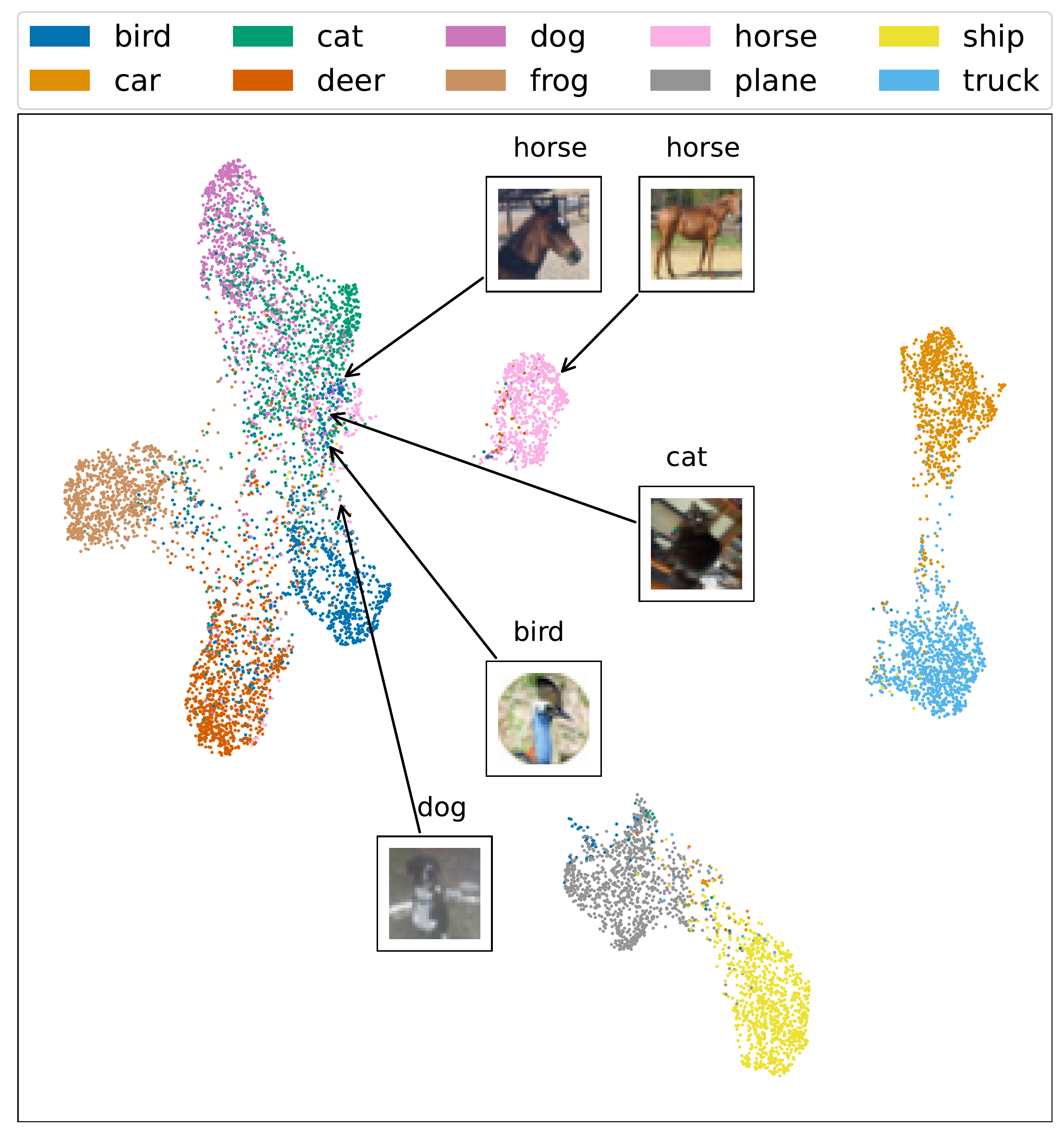}
		\label{fig:umap visualization clustering TSP ViT 1 CIFAR10 pretrained on CIFAR 10}
	}
	\hfill
\subfigure[Source dataset: ImageNet-1k]{	\includegraphics[height=0.4 \linewidth]{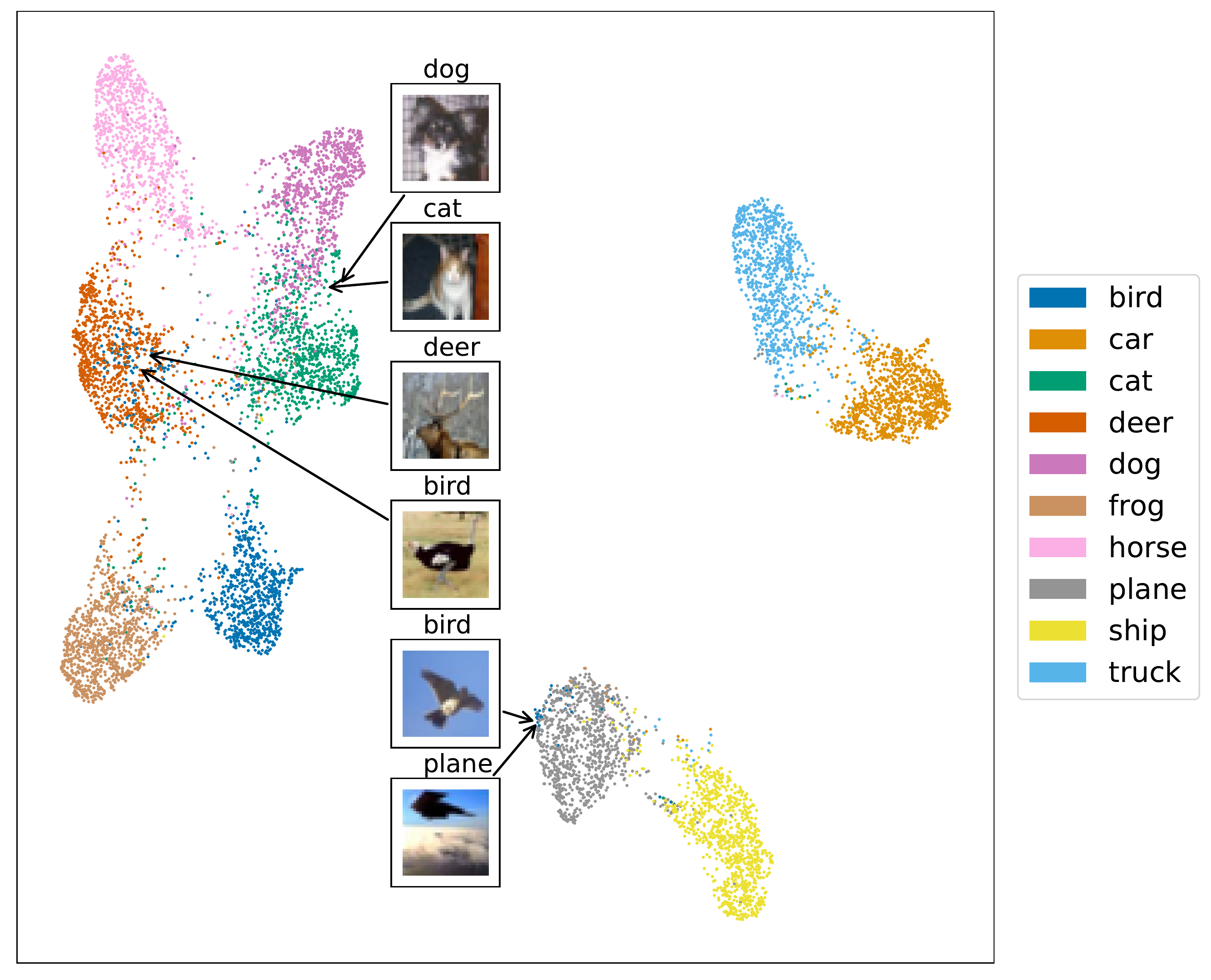}
	\label{fig:umap visualization clustering TSP CIFAR10}
}

	\subfigure[Source dataset: CIFAR10]{
	\includegraphics[height=0.4\linewidth]{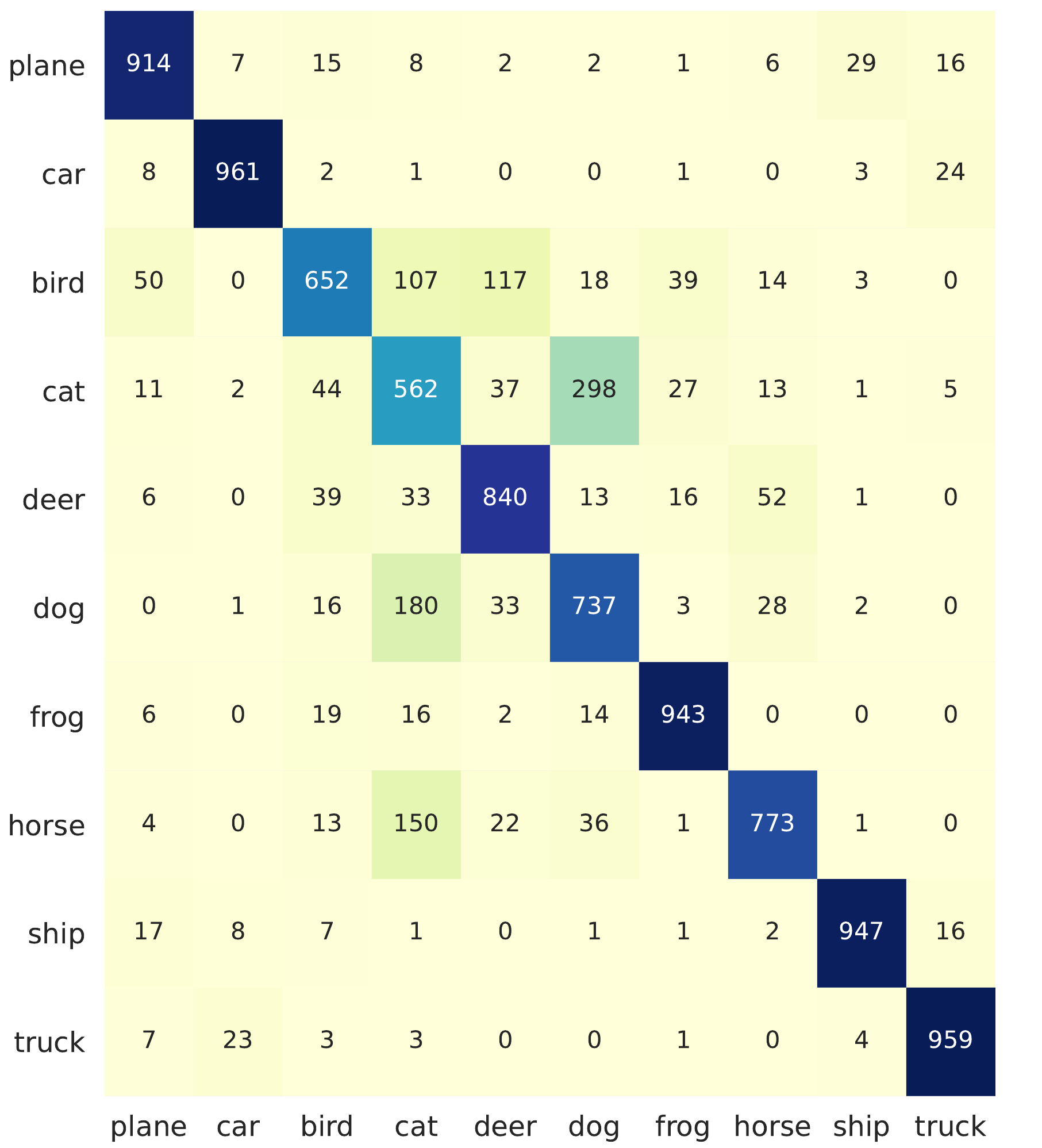}
	\label{fig:confusion matrix clustering TSP ViT 1 CIFAR10 pretrained on CIFAR 10}
}
\hfill
\subfigure[Source dataset: ImageNet-1k]{	\includegraphics[height=0.4 \linewidth]{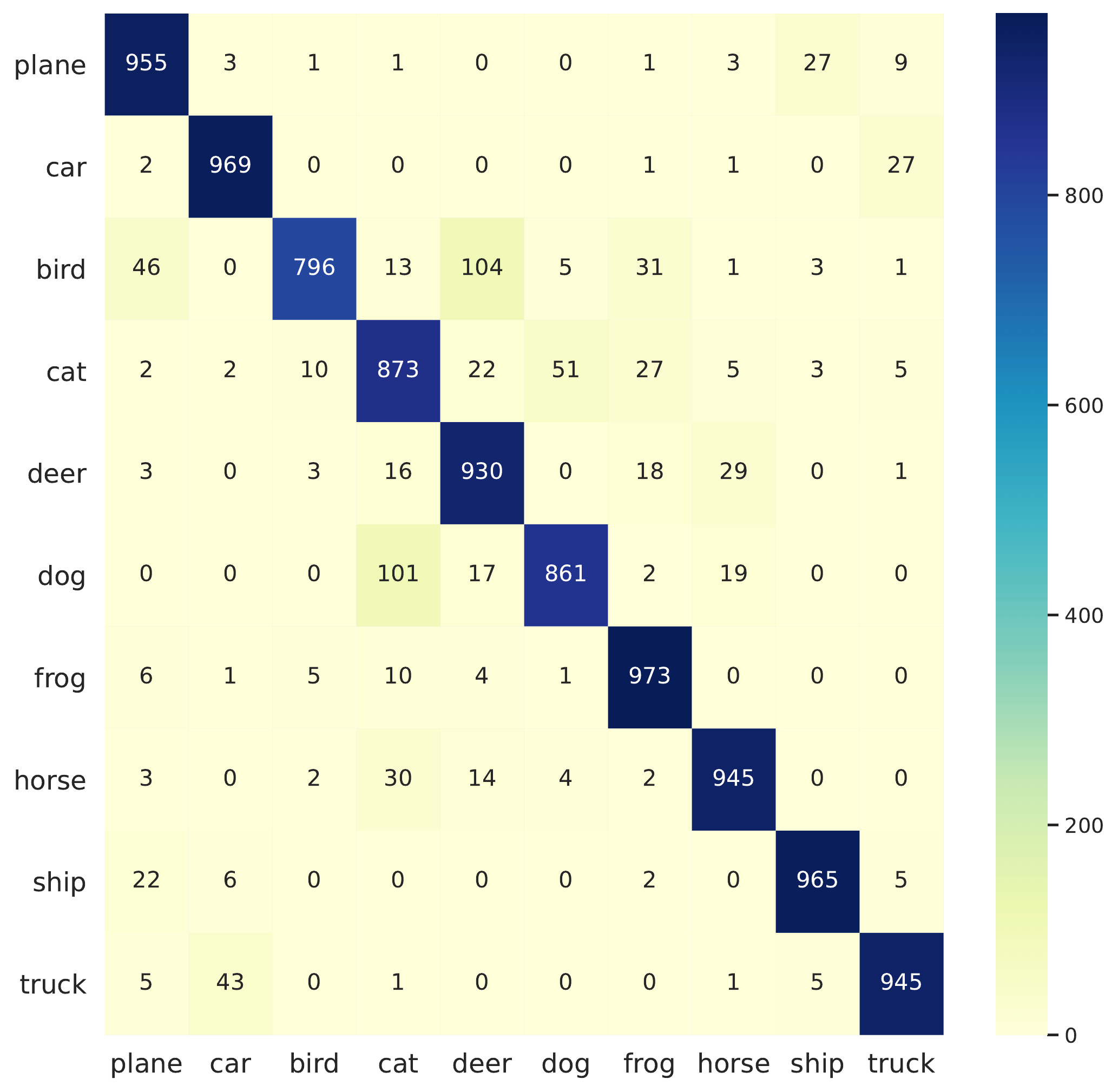}
	\label{confusion matrix clustering TSP CIFAR10}
}

	\caption{Distribution and confusion matrix of the CIFAR-10 data points after the
		clustering head of TSP-ViT1 with different source datasets.  }
	\label{figure:umap visualization clustering  CIFAR10}
\end{figure}

The distributions and confusion matrices of the clustering results on CIFAR 10 with TSP-ViT1 on different setting are shown in Figure \ref{figure:umap visualization clustering  CIFAR10}. The clustering result of TSP-ViT1 with source dataset ImageNet-1k performs better than that with source dataset ImageNet-1k, which is consistent to the representation performance of pretraining shown in Figure \ref{figure:umap visualization ViTs pretrained on different source datasets}. The good performance of TSP-ViT1 pretrained by ImageNet-1k is due to that the large quantities of data empowers the clustering structure preservation on the ViT architectures. That is, in BaW strategy, large group of white sheet can lead to a superior separation between black sheep.

%\section{Appendix}
%
%
%Optionally include extra information (complete proofs, additional experiments and plots) in the appendix.
%This section will often be part of the supplemental material.

\end{document}